%% file: main.tex
\documentclass[runningheads]{llncs}
\usepackage{graphicx}

\usepackage{tikz}
\usepackage{comment}
\usepackage{amsmath,amssymb} %
\usepackage{color}

\usepackage[accsupp]{axessibility}  %

\usepackage{multirow}
\usepackage{booktabs}
\usepackage{threeparttable} %
\usepackage{blindtext}
\usepackage{adjustbox}
\usepackage{footmisc} %
\usepackage{diagbox}
\usepackage[pagebackref=true,breaklinks=true,colorlinks,bookmarks=false]{hyperref}

\newcommand\ours{{{\mbox{Donut}}}\xspace}
\newcommand\oursb{{\textbf{\mbox{Donut}}}\xspace}

\usepackage[width=122mm,left=12mm,paperwidth=146mm,height=193mm,top=12mm,paperheight=217mm]{geometry}

\begin{document}
\pagestyle{headings}
\mainmatter

\title{OCR-free Document Understanding Transformer}

\title{OCR-free Document Understanding Transformer}
\author{Geewook Kim$^{1}$\thanks{\ Corresponding author: gwkim.rsrch@gmail.com}, Teakgyu Hong$^{4}$\thanks{\ This work was done while the authors were at NAVER CLOVA.}, Moonbin Yim$^{2}$$^{\dag}$, Jeongyeon Nam$^{1}$, \\ Jinyoung Park$^{5}$$^{\dag}$, Jinyeong Yim$^{6}$$^{\dag}$, Wonseok Hwang$^{7}$$^{\dag}$, Sangdoo Yun$^{3}$, \\ Dongyoon Han$^{3}$, \and Seunghyun Park$^{1}$}
\authorrunning{G. Kim et al.}
\institute{$^{1}$NAVER CLOVA \qquad $^{2}$NAVER Search \qquad $^{3}$NAVER AI Lab \\ $^{4}$Upstage\qquad $^{5}$Tmax\qquad $^{6}$Google\qquad $^{7}$LBox}
\maketitle

\begin{abstract}
\input{tex/0.abstract}
\end{abstract}

\section{Introduction}
\input{tex/1.introduction}

\section{Method}
\subsection{Preliminary: background}
\input{tex/2.background}

\input{tex/3.method}

\section{Experiments and Analyses}\label{sec:exp}
\input{tex/4.experiments}

\section{Related Work}
\input{tex/5.related}

\section{Conclusions} %
\input{tex/6.conclusion}

\clearpage

\bibliographystyle{splncs04}

\input{main.bbl}
\clearpage

\appendix
\renewcommand{\thesection}{\Alph{section}}
\renewcommand{\thesubsection}{\Alph{section}.\arabic{subsection}}
\setcounter{figure}{0}
\renewcommand{\thefigure}{\Alph{figure}}
\section{Appendix}

\subsection{Details of OCR Engines (MS, CLOVA, Easy, Paddle)}\label{sec:detail_of_ocr_engines}
Current state-of-the-art visual document understanding (VDU) backbones, such as BROS~\cite{hong2021bros}, LayoutLM~\cite{xu2019_layoutLM} and LayoutLMv2~\cite{xu-etal-2021-layoutlmv2}, are dependent on off-the-shelf OCR engines.
These backbones take the output of OCR as their (one of) input features.
For the OCR-dependent methods, in our experiments, we use state-of-the-art OCR engines that are publicly available, including 2 OCR API products (i.e., MS OCR\footnote{\url{https://docs.microsoft.com/en-us/azure/cognitive-services/computer-vision/overview-ocr}.\label{footnote_ms_url}} and CLOVA OCR\footnote{\url{https://clova.ai/ocr/en}.\label{footnote_clova_url}}) and 2 open-source OCR models (i.e., Easy OCR\footnote{\url{https://github.com/JaidedAI/EasyOCR}.\label{footnote_easy_url}} and Paddle OCR\footnote{\url{https://github.com/PaddlePaddle/PaddleOCR}.\label{footnote_paddle_url}}).
In the main paper, Paddle OCR is used for the Chinese train ticket dataset~\cite{eaten} and CLOVA OCR is used for the rest datasets in the document information extraction (IE) tasks. 
MS OCR is used to measure the running time of the LayoutLM family in document classification and visual question answering (VQA) tasks, following the previous work of Xu et al.~\cite{xu-etal-2021-layoutlmv2}.
Each OCR engine is explained in the following.

\subsubsection{MS OCR}
MS OCR\footref{footnote_ms_url} is the latest OCR API product from Microsoft and used in several recent VDU methods, e.g., LayoutLMv2~\cite{xu-etal-2021-layoutlmv2}.
This engine supports 164 languages for printed text and 9 languages for handwritten text (until 2022/03).

\subsubsection{CLOVA OCR}
CLOVA OCR\footref{footnote_clova_url} is an API product from NAVER CLOVA and is specialized in document IE tasks.
This engine supports English, Japanese and Korean (until 2022/03).
In the ablation experiments on the CORD dataset~\cite{park2019cord} (Figure 9 in the main paper), the CLOVA OCR achieved the best accuracy.

\subsubsection{Easy OCR}
Easy OCR\footref{footnote_easy_url} is a ready-to-use OCR engine that is publicly available at GitHub. This engine supports more than 80 languages (until 2022/03). Unlike the aforementioned two OCR products (i.e., MS OCR and CLOVA OCR),
this engine is publicly opened and downloadable.\footref{footnote_easy_url}
The entire model architecture is based on the modern deep-learning-based OCR modules~\cite{baek2019craft,baek2019wrong} with some modifications to make the model lighter and faster. The total number of model parameters is 27M which is small compared to the state-of-the-art models~\cite{baek2019craft,baek2019wrong}.

\subsubsection{Paddle OCR}
Paddle OCR\footref{footnote_paddle_url} is an open-source OCR engine available at GitHub. We used a lightweight (i.e., mobile) version of the model which is specially designed for a fast and light OCR of English and Chinese texts. The model is served on a CPU environment and the size of the model is extremely small, which is approximately 10M.

\begin{figure}[t!]
    \centering
    \includegraphics[width=\linewidth]{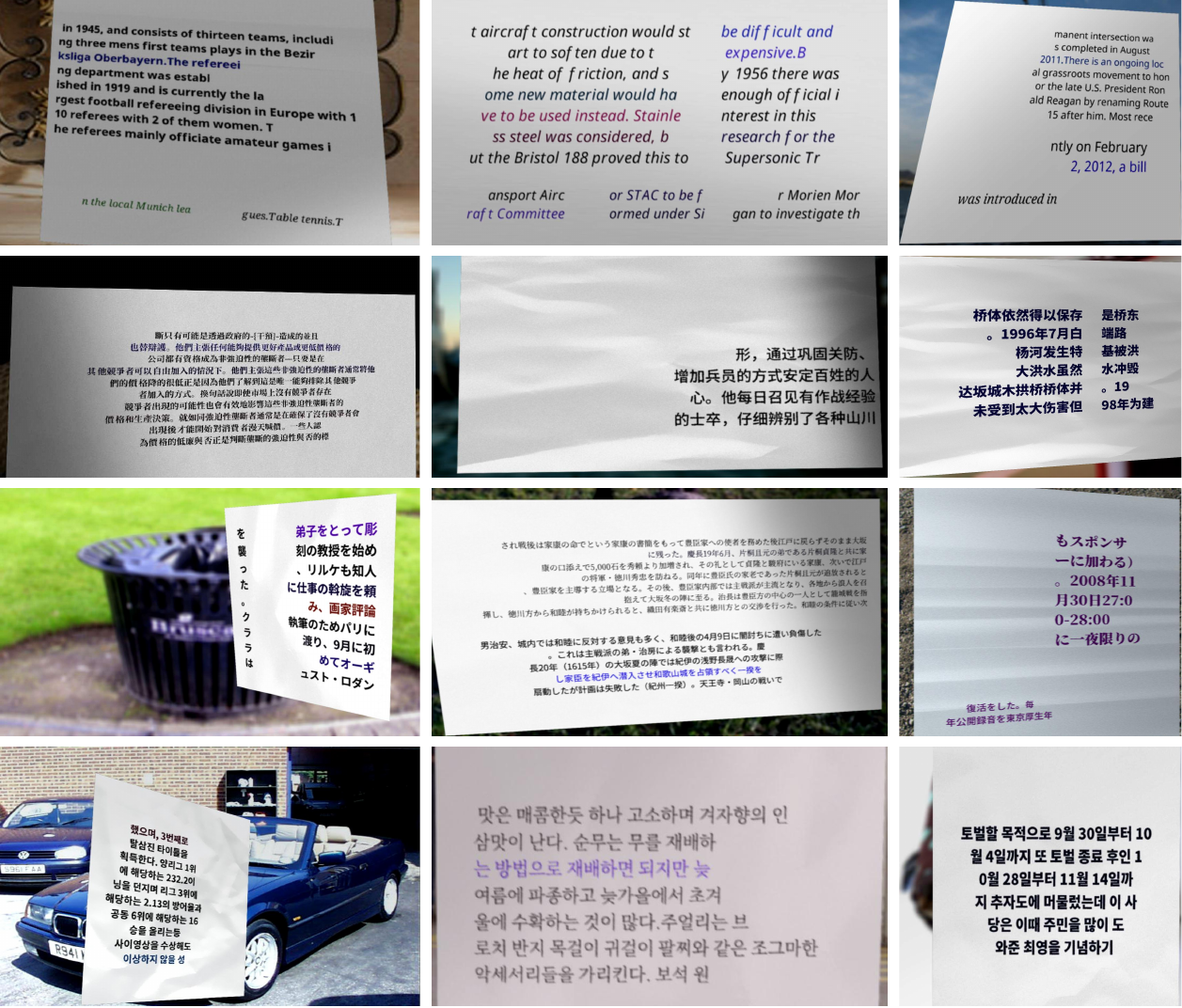}
    \caption{{\bf Examples of SynthDoG.} English, Chinese, Japanese and Korean samples are shown (from top to bottom). Although the idea is simple, these synthetic samples play an important role in the pre-training of \ours. Please, see Figure 7 in the main paper for details}
    \label{fig:more_synthdog}
\end{figure}

\subsection{Details of Synthetic Document Generator (SynthDoG)}\label{sec:detail_of_synthdog}
In this section, we explain the components of the proposed Synthetic Document Generator (SynthDoG) in detail. 
The entire pipeline basically follows Yim et al.~\cite{synthtiger}.
Our source code is available at \url{https://github.com/clovaai/donut}.
More samples are shown in Figure~\ref{fig:more_synthdog}.

\subsubsection{Background}
Background images are sampled from ImageNet~\cite{deng2009imagenet}.
Gaussian blur is randomly applied to the background image to represent out-of-focus effects.

\subsubsection{Document}
Paper textures are sampled from the photos that we collected. The texture is applied to an white background. In order to make the texture realistic, random elastic distortion and Gaussian noise are applied. To represent various view angles in photographs, a random perspective transformation is applied to the image.

\subsubsection{Text Layout and Pattern}
To mimic the layouts in real-world documents, a heuristic rule-based pattern generator is applied to the document image region to generate text regions. The main idea is to set multiple squared regions to represent text paragraphs. 
Each squared text region is then interpreted as multiple lines of text. The size of texts and text region margins are chosen randomly.

\subsubsection{Text Content and Style}
We prepare the multi-lingual text corpora from Wikipedia.\footnote{\url{https://dumps.wikimedia.org}.}
We use Noto fonts\footnote{\url{https://fonts.google.com/noto}.} since it supports various languages. SynthDoG samples texts and fonts from these resources and the sampled texts are rendered in the regions that are generated by the layout pattern generator. The text colors are randomly assigned.

\subsubsection{Post-processing}
Finally, some post-processing techniques are applied to the output image. In this process, the color, brightness, and contrast of the image are adjusted. In addition, shadow effect, motion blur, Gaussian blur, and JPEG compression are applied to the image.

\subsection{Details of Document Information Extraction} %
Information Extraction (IE) on documents is an arduous task since it requires (a) reading texts, (b) understanding the meaning of the texts, and (c) predicting the relations and structures among the extracted information. Some previous works have only focused on extracting several pre-defined key information~\cite{eaten}. In that case, only (a) and (b) are required for IE models. We go beyond the previous works by considering (c) also. Although the task is complex, its interface (i.e., the format of input and output) is simple. 
In this section, for explanation purposes, we show some sample images (which are the raw input of the IE pipeline) with the output of Donut.

\begin{figure}[t!]
    \centering
    \includegraphics[width=\linewidth]{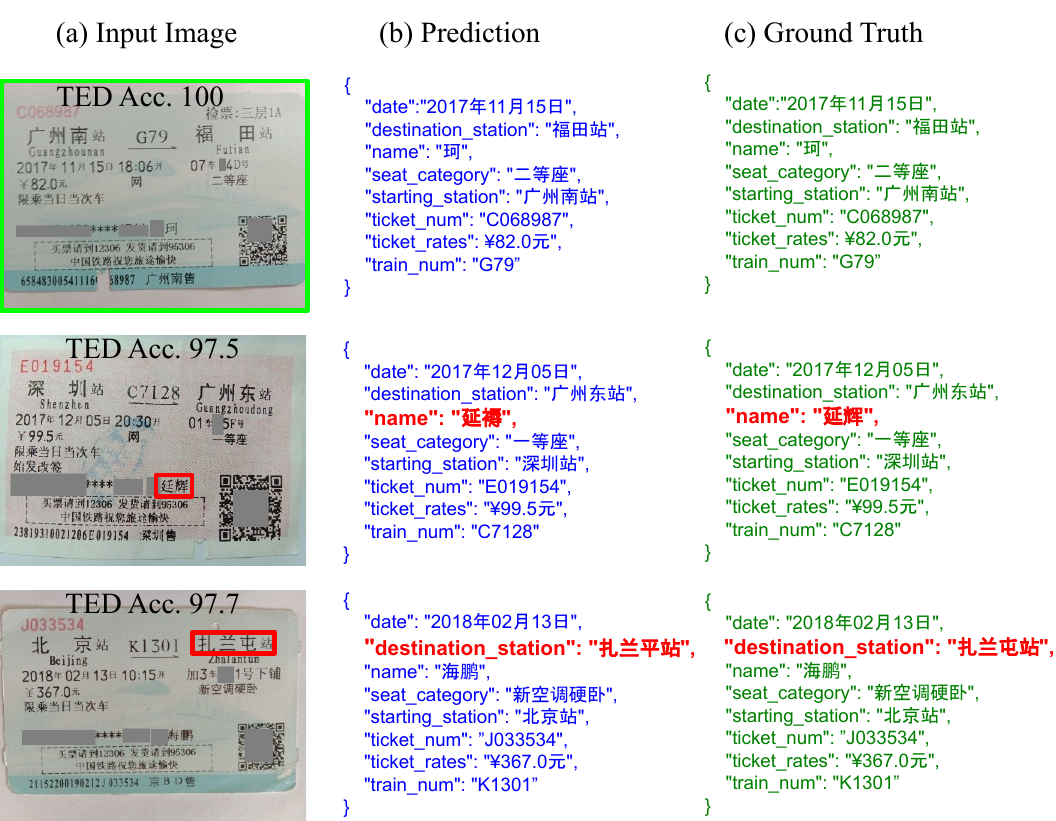}
    \caption{{\bf Examples of \textit{Ticket}~\cite{eaten} with \ours predictions.} There is no hierarchy in the structure of information (i.e., depth $=1$) and the location of each key information is almost fixed. Failed predictions are marked and bolded (red)}
    \label{fig:ticket_example}
\end{figure}

\begin{figure}[t!]
    \centering
    \includegraphics[width=\linewidth]{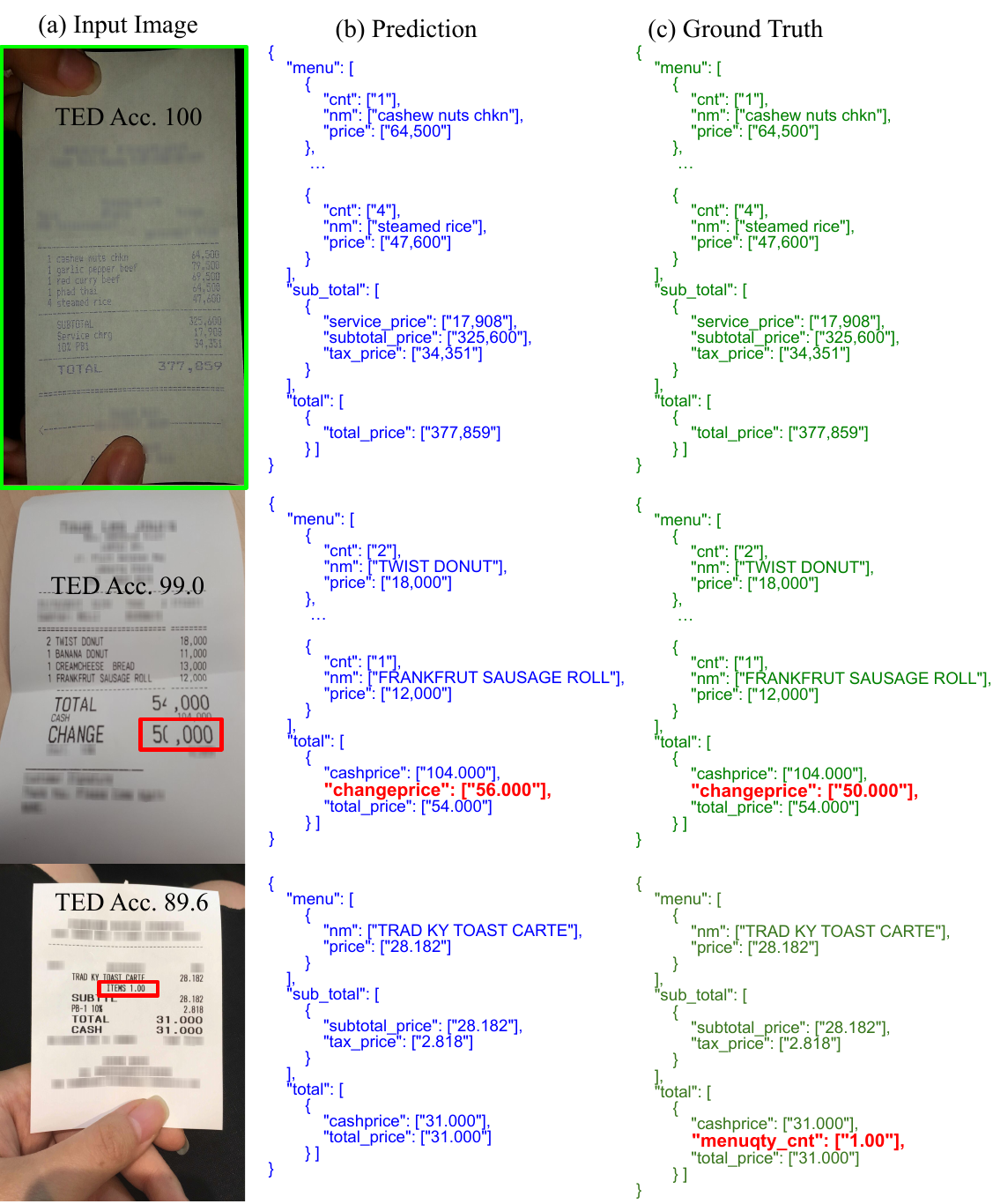}
    \caption{{\bf Examples of \textit{CORD}~\cite{park2019cord} with \ours predictions.} There is a hierarchy in the structure of information (i.e., depth $=2$). \oursb not only reads some important key information from the image, but also predicts the relationship among the extracted information (e.g., the name, price, and quantity of each menu item are grouped)}
    \label{fig:cord_example}
\end{figure}

\begin{figure}[t!]
    \centering
    \includegraphics[width=\linewidth]{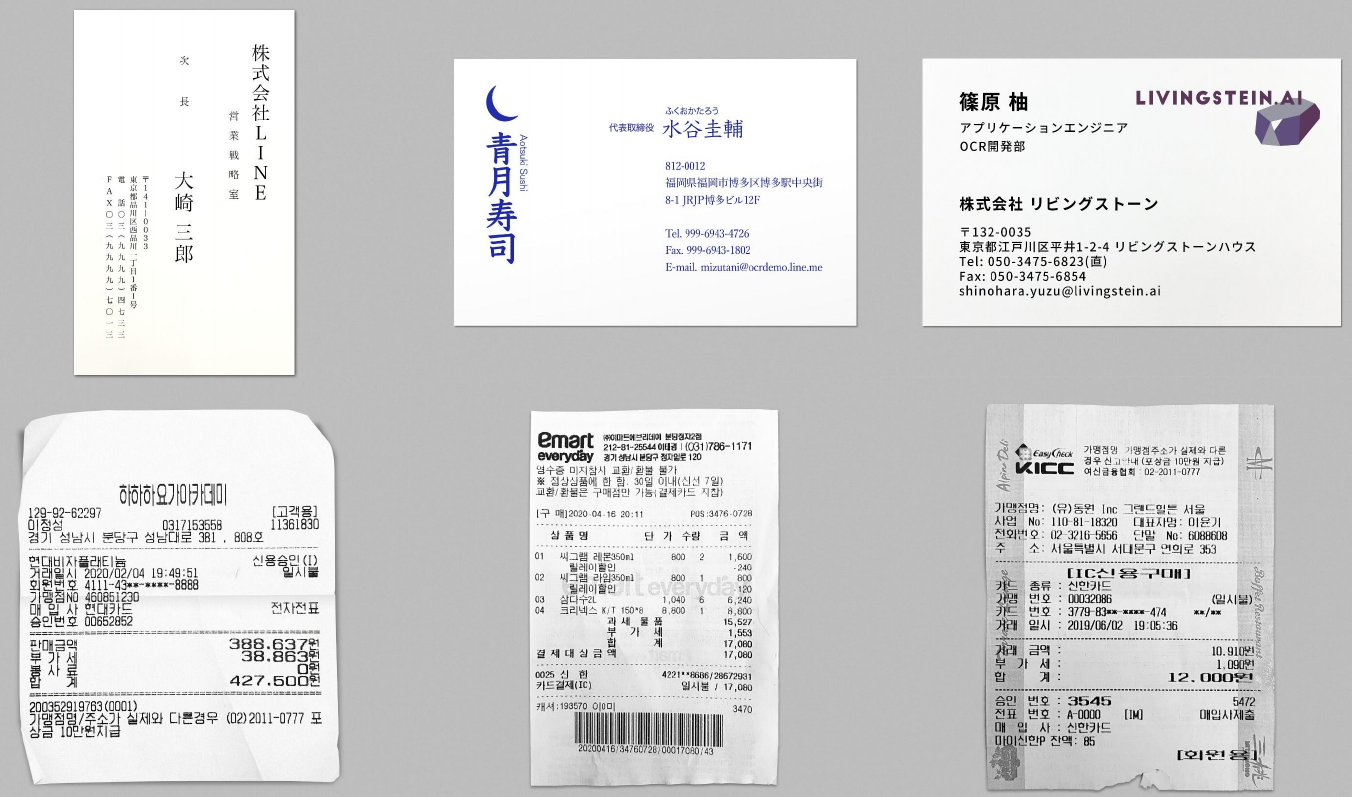}
    \caption{{\bf Examples of \textit{Business Card} (top) and \textit{Receipt} (bottom).} Due to strict industrial policies on the private industrial datasets from our active products, real-like high-quality samples are shown instead}
    \label{fig:kor_jpn_example}
\end{figure}

In the main paper, we test four datasets including two public benchmarks (i.e., \textit{CORD}~\cite{park2019cord} and \textit{Ticket}~\cite{eaten}) and two private industrial datasets (i.e., \textit{Business Card} and \textit{Receipt}). 
Figure~\ref{fig:ticket_example} shows examples of \textit{Ticket} with the outputs of Donut.
Figure~\ref{fig:cord_example} shows examples of \textit{CORD} with the outputs of Donut.
Due to strict industrial policies on the private industrial datasets, we instead show some real-like high-quality samples of \textit{Business Card} and \textit{Receipt} in Figure~\ref{fig:kor_jpn_example}.

\begin{figure}[t]
  \centering
  \includegraphics[width=\linewidth]{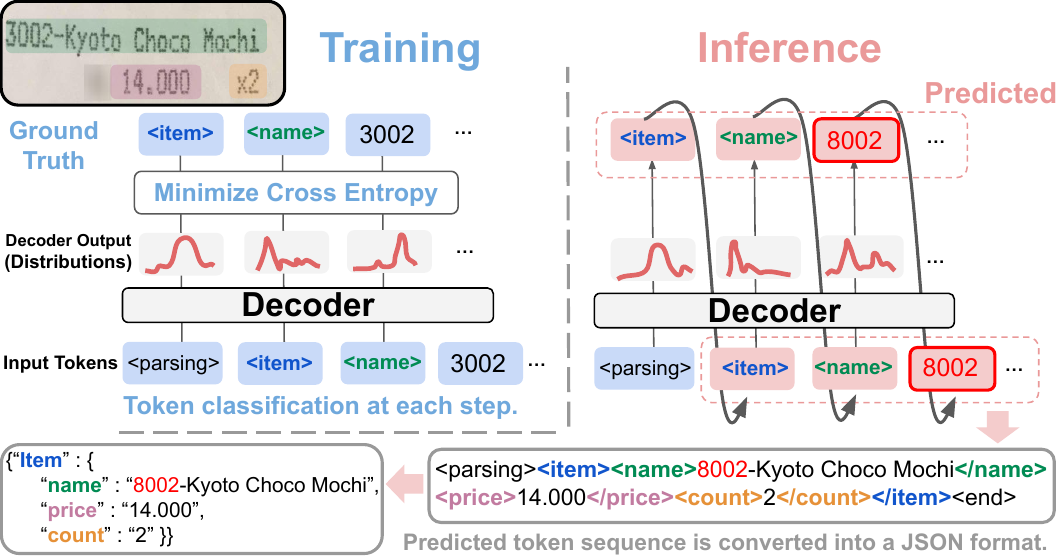}
  \vspace{-6.8995mm}
  \caption{\textbf{Donut training scheme with teacher forcing and decoder output format examples.} The model is trained to minimize cross-entropy loss of the token classifications simultaneously. At inference, the predicted token from the last step is fed to the next} %
  \label{fig:onecol}
\end{figure}

\subsection{Details of Model Training Scheme and Output Format}\label{sec:detail_of_scheme_and_format}
In the model architecture and training objective, we basically followed the original Transformer~\cite{vaswani2017transformer}, which uses a Transformer encoder-decoder architecture and a teacher-forcing training scheme.
The teacher-forcing scheme is a model training strategy that uses the ground truth as input instead of model output from a previous time step.
Figure~\ref{fig:onecol} shows a details of the model training scheme and decoder output format.

\subsection{Implementation and Training Hyperparameters}\label{sec:detail_of_implementation_and_hyperparams}
The codebase and settings are available at GitHub.\footnote{\url{https://github.com/clovaai/donut}.}
We implement the entire model pipeline with Huggingface's \texttt{transformers}\footnote{\url{https://github.com/huggingface/transformers}.}~\cite{wolf-etal-2020-transformers} and an open-source library \texttt{TIMM} (PyTorch image models)\footnote{\url{https://github.com/rwightman/pytorch-image-models}.}~\cite{rw2019timm}.

For all model training, we use a half-precision (fp16) training.
We train \ours using Adam optimizer~\cite{Adamoptim} by decreasing the learning rate as the training progresses.
The initial learning rate of pre-training is set to 1e-4 and that of fine-tuning is selected from 1e-5 to 1e-4.
We pre-train the model for 200K steps with 64 NVIDIA A100 GPUs and a mini-batch size of 196,
which takes about 2-3 GPU days.
We also apply a gradient clipping technique where a maximum gradient norm is selected from 0.05 to 1.0.
The input resolution of \ours is set to 2560$\times$1920 at the pre-training phase.
In downstream tasks, the input resolutions are controlled.
In some downstream document IE experiments, such as, \textit{CORD}~\cite{park2019cord}, \textit{Ticket}~\cite{eaten} and \textit{Business Card}, smaller size of input resolution, e.g., 1280$\times$960, is tested.
With the 1280$\times$960 setting, the model training cost of \ours was small.
For example, the model fine-tuning on \textit{CORD} or \textit{Ticket} took approximately 0.5 hours with one A100 GPU. However, when we set the 2560$\times$1920 setting for larger datasets, e.g., \textit{RVL-CDIP} or \textit{DocVQA}, the cost increased rapidly. With 64 A100 GPUs, \textit{DocVQA} requires one GPU day and \textit{RVL-CDIP} requires two GPU days approximately. This is not surprising in that increasing the input size for a precise result incurs higher computational costs in general. Using an efficient attention mechanism~\cite{wang2020linformer} may avoid the problem in architectural design, but we use the original Transformer~\cite{vaswani2017transformer} as we aim to present a simpler architecture in this work. Our preliminary experiments in smaller resources are available in Appendix~\ref{sec:smaller_resources}.

For the implementation of document IE baselines, we use the \texttt{transformers} library for BERT~\cite{devlinBERT2018}, BROS~\cite{hong2021bros}, LayoutLMv2~\cite{xu-etal-2021-layoutlmv2,layoutxlm} and WYVERN~\cite{hwang2021costeffective}.
For the SPADE~\cite{hwang-etal-2021-spatial} baseline, the official implementation\footnote{\url{https://github.com/clovaai/spade}.} is used.
The models are trained using NVIDIA P40, V100, or A100 GPUs. The major hyperparameters, such as initial learning rate and number of epochs, are adjusted by monitoring the scores on the validation set. The architectural details of the OCR-dependent VDU backbone baselines (e.g., LayoutLM and LayoutLMv2) are available in Appendix~\ref{sec:detail_of_VDU_backbone}.

\subsection{Preliminary Experiments in Smaller Resources}\label{sec:smaller_resources}
In our preliminary experiments, we pre-trained \ours with smaller resources (denoted as Donut$_{\text{Proto}}$), i.e., smaller data (SynthDoG 1.2M) and fewer GPUs (8 V100 GPUs for 5 days). The input size was 2048$\times$1536. In this setting, Donut$_{\text{Proto}}$ also achieved comparable results on \textit{RVL-CDIP} and \textit{CORD}. The accuracy on \textit{RVL-CDIP} was 94.5 and \textit{CORD} was 85.4.
After the preliminaries, we have scaled the model training with more data.

\subsection{Details of OCR-dependent Baseline Models}\label{sec:detail_of_VDU_backbone}
In this section, we provide a gentle introduction to the general-purpose VDU backbones, such as LayoutLM~\cite{xu2019_layoutLM} and LayoutLMv2~\cite{xu-etal-2021-layoutlmv2}.
To be specific, we explain how the conventional backbones perform downstream VDU tasks; document classification, IE, and VQA.
Common to all tasks, the output of the OCR engine is used as input features of the backbone.
That is, the extracted texts are sorted and converted to a sequence of text tokens. The sequence is passed to the Transformer encoder to get contextualized output vectors.
The vectors are used to get the desired output.
The difference in each task depends on a slight modification on the input sequence or on the utilization of the output vectors.

\subsubsection{Document Classification}
At the start of the input sequence, a special token \texttt{[CLS]} is appended.
The sequence is passed to the backbone to get the output vectors.
With a linear mapping and softmax operation, the output vector of the special token \texttt{[CLS]} is used to get a \textit{class-label} prediction.

\subsubsection{Document IE}
With a linear mapping and softmax operation, the output vector sequence is converted to a \textit{BIO-tag} sequence~\cite{hwang2019pot}.

\paragraph{IE on 1-depth structured documents}
When there is no hierarchical structure in the document (See Figure~\ref{fig:ticket_example}), the tag set is defined as \{``B$_{k}$'', ``I$_{k}$'', ``O''  $\mid k\in$ pre-defined keys\}.
``B$_{k}$'' and ``I$_{k}$'' are tags that represent the beginning (B) and the inside (I) token of the key $k$ respectively.
The ``O'' tag indicates that the token belongs to no key information. 

\paragraph{IE on $n$-depth structured documents}
When there are hierarchies in the structure (See Figure~\ref{fig:cord_example}), the BIO-tags are defined for each hierarchy level.
In this section, we explain a case where the depth of structure is $n=2$.
The tag set is defined as \{``B$_{g}$.B$_{k}$'', ``B$_{g}$.I$_{k}$'', ``I$_{g}$.B$_{k}$'', ``I$_{g}$.I$_{k}$'', ``O''  $\mid g\in$ pre-defined parent keys, $k\in$ pre-defined child keys\}.
For instance, the Figure~\ref{fig:cord_example} shows an example where a parent key is ``menu'' and related child keys are \{``cnt'', ``nm'', ``price''\}.
``B$_{g}$'' represents that one group (i.e., a parent key such as ``menu'') starts, and ``I$_{g}$'' represents that the group is continuing. 
Separately from the BI tags of the parent key (i.e., ``B$_{g}$'' and ``I$_{g}$''), the BI tags of each child key (i.e., ``B$_{k}$'' and ``I$_{k}$'') work the same as in the case of $n=1$.
This BIO-tagging method is also known as \textit{Group BIO-tagging} and the details are also available in Hwang et al.~\cite{hwang2019pot}.

\subsubsection{Document VQA}
With a linear mapping and softmax operation, the output vector sequence is converted to a \textit{span-tag} sequence.
For the input token sequence, the model finds the beginning and the end of the answer span.
Details can also be found in the Section 4.2 of Devlin et al.~\cite{devlinBERT2018}.

\end{document}

%% file: tex/0.abstract.tex
Understanding document images (\textit{e.g.}, invoices) is a core but challenging task since it requires complex functions such as \textit{reading text} and a \textit{holistic understanding of the document}.
Current Visual Document Understanding (VDU) methods outsource the task of reading text to off-the-shelf Optical Character Recognition (OCR) engines and focus on the understanding task with the OCR outputs.
Although such OCR-based approaches have shown promising performance, they suffer from 1) high computational costs for using OCR; 2) inflexibility of OCR models on languages or types of documents; 3) OCR error propagation to the subsequent process. 
To address these issues, in this paper, we introduce a novel OCR-free VDU model named \oursb, which stands for \textbf{Do}cume\textbf{n}t \textbf{u}nderstanding \textbf{t}ransformer.
As the first step in OCR-free VDU research, we propose a simple architecture (\textit{i.e.}, Transformer) with a pre-training objective (\textit{i.e.,} cross-entropy loss).
Donut is conceptually simple yet effective. Through extensive experiments and analyses, we show a simple OCR-free VDU model, \ours, achieves state-of-the-art performances on various VDU tasks in terms of both speed and accuracy.
In addition, we offer a synthetic data generator that helps the model pre-training to be flexible in various languages and domains.
The code, trained model, and synthetic data are available at \url{https://github.com/clovaai/donut}.

%% file: tex/1.introduction.tex
\begin{figure*}[t]
\centering
  \includegraphics[width=\textwidth]{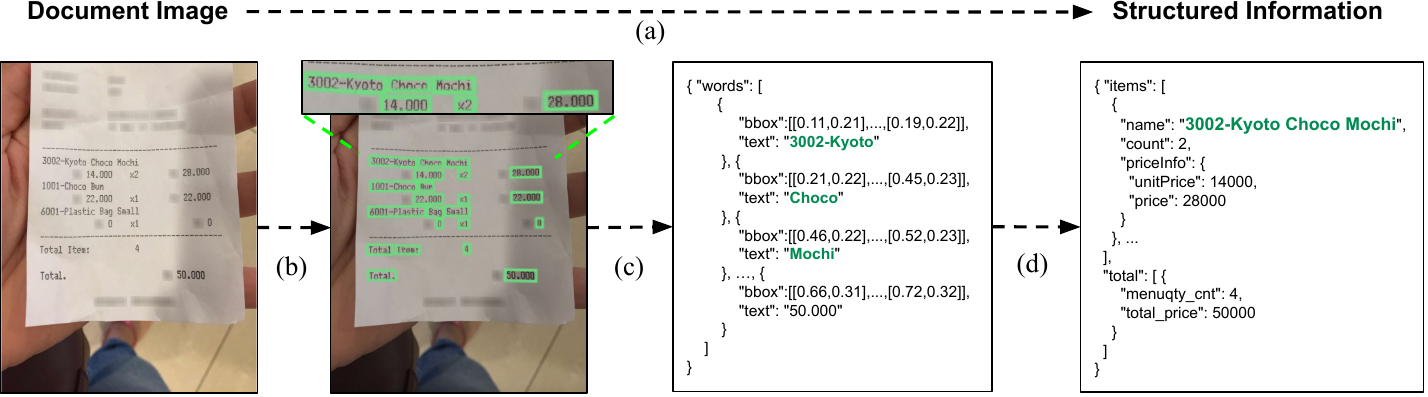} 
  \caption{{\bf The schema of the conventional document information extraction (IE) pipeline.} (a) The goal is to extract the structured information from a given semi-structured document image. In the pipeline, (b) text detection is conducted to obtain text locations and (c) each box is passed to the recognizer to comprehend characters. (d) Finally, the recognized texts and its locations are passed to the following module to be processed for the desired structured form of the information} \label{fig:problem_definition}
\end{figure*}

\begin{figure}[t]
\begin{center}
\includegraphics[width=\linewidth]{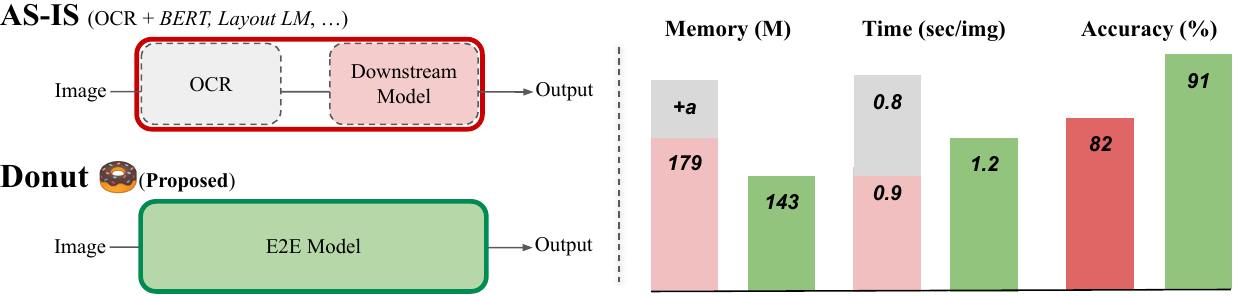} 
\end{center}
\vspace*{-0.4cm}
\hspace*{1.4cm} {\scriptsize (a) Pipeline Overview.} \hspace*{3.0cm} {\scriptsize (b) System Benchmarks.}
\vspace*{-0.2cm}
\caption{\textbf{The pipeline overview and benchmarks.} The proposed end-to-end model, \oursb, outperforms the recent OCR-dependent VDU models in memory, time cost and accuracy. Performances on visual document IE~\cite{park2019cord} are shown in (b). More results on various VDU tasks are available at Section~\ref{sec:exp} showing the same trend} \label{fig:teaser_summary}
\end{figure}

Document images, such as commercial invoices, receipts, and business cards, are easy to find in modern working environments.
To extract useful information from such document images, Visual Document Understanding (VDU) has not been only an essential task for industry, but also a challenging topic for researchers, with applications including document classification~\cite{Kang2014ConvolutionalNN,7333933}, information extraction~\cite{hwang2019pot,majumder2020representation}, and visual question answering~\cite{mathew2021docvqa,icdar21docvqa}.

Current VDU methods~\cite{hwang2019pot,hwang2020spade,xu2019_layoutLM,xu-etal-2021-layoutlmv2,hong2021bros} solve the task in a two-stage manner: 1) reading the texts in the document image; 2) holistic understanding of the document.
They usually rely on deep-learning-based Optical Character Recognition (OCR)~\cite{baek2019craft,baek2019wrong} for the text reading task and focus on modeling the understanding part. 
For example, as shown in Figure~\ref{fig:problem_definition}, a conventional pipeline for extracting structured information from documents (a.k.a. document parsing) consists of three separate modules for text detection, text recognition, and parsing~\cite{hwang2019pot,hwang2020spade}.

However, the OCR-dependent approach has critical problems.
First of all, using OCR as a pre-processing method is expensive.
We can utilize pre-trained off-the-shelf OCR engines; however, the computational cost for inference would be expensive for high-quality OCR results. 
Moreover, the off-the-shelf OCR methods rarely have flexibility dealing with different languages or domain changes, which may lead to poor generalization ability. 
If we train an OCR model, it also requires extensive training costs and large-scale datasets~\cite{baek2019craft,baek2019wrong,Liu_2020_CVPR,spts}.
Another problem is, OCR errors would propagate to the VDU system and negatively influence subsequent processes~\cite{ocr_error_negative,hwang2021costeffective}. This problem becomes more severe in languages with complex character sets, such as Korean or Chinese, where the quality of OCR is relatively low~\cite{rijhwani-etal-2020-ocr}.
To deal with this, post-OCR correction module~\cite{schaefer-neudecker-2020-two,rijhwani-etal-2020-ocr,duong-etal-2021-unsupervised} is usually adopted. However, it is not a practical solution for real application environments since it increases the entire system size and maintenance cost.

We go beyond the traditional framework by modeling a direct mapping from a raw input image to the desired output without OCR.
We introduce a new OCR-free VDU model to address the problems induced by the OCR-dependency.
Our model is based on Transformer-only architecture, referred to as \textbf{Do}cume\textbf{n}t \textbf{u}nderstanding \textbf{t}ransformer (\oursb), following the huge success in vision and language~\cite{devlinBERT2018,dosovitskiy2020vit,pmlr-v139-kim21k}.
We present a minimal baseline including a simple architecture and pre-training method. 
Despite its simplicity, \oursb shows comparable or better overall performance than previous methods as shown in Figure~\ref{fig:teaser_summary}. 

We take pre-train-and-fine-tune scheme~\cite{devlinBERT2018,xu2019_layoutLM} on \oursb training. 
In the pre-training phase, \oursb learns \textit{how to read the texts} by predicting the next words by conditioning jointly on the image and previous text contexts. 
\oursb is pre-trained with document images and their text annotations. 
Since our pre-training objective is simple (\textit{i.e.}, reading the texts), we can realize domain and language flexibility straightforwardly pre-training with synthetic data.
During fine-tuning stage, \oursb learns \textit{how to understand the whole document} according to the downstream task. 
We demonstrate \oursb has a strong understanding ability through extensive evaluation on various VDU tasks and datasets. 
The experiments show a simple OCR-free VDU model can achieve state-of-the-art performance in terms of both speed and accuracy. 

The contributions are summarized as follows:
\begin{enumerate}
\item[1.] We propose a novel OCR-free approach for VDU. To the best of our knowledge, this is the first method based on an OCR-free Transformer trained in end-to-end manner.
\item[2.] We introduce a simple pre-training scheme that enables the utilization of synthetic data. By using our generator {SynthDoG}, we show \oursb can easily be extended to a multi-lingual setting, which is not applicable for the conventional approaches that need to retrain an off-the-shelf OCR engine. 
\item[3.] We conduct extensive experiments and analyses on both public benchmarks and private industrial datasets, showing that the proposed method achieves not only state-of-the-art performances on benchmarks but also has many practical advantages (e.g., \textit{cost-effective}) in real-world applications. 
\item[4.] The codebase, pre-trained model, and synthetic data are available at GitHub.\footnote{\url{https://github.com/clovaai/donut}\label{footnote_code}.}
\end{enumerate}

%% file: tex/2.background.tex
There have been various visual document understanding (VDU) methods to understand and extract essential information from the semi-structured documents such as receipts~\cite{8977955,hwang-etal-2021-spatial,hong2021bros}, invoices~\cite{8978079}, and form documents~\cite{7333829,8977962,majumder-etal-2020-representation}. 

Earlier VDU attempts have been done with OCR-independent visual backbones~\cite{Kang2014ConvolutionalNN,7333933,7333910,eaten,docreader}, but the performances are limited. %
Later, with the remarkable advances of OCR~\cite{baek2019craft,baek2019wrong} and BERT~\cite{devlinBERT2018}, various OCR-dependent VDU models have been proposed by combining them~\cite{hwang2019pot,hwang2020spade,hwang2021costeffective}. More recently, in order to get a more general VDU, most state-of-the-arts~\cite{xu-etal-2021-layoutlmv2,hong2021bros} use both powerful OCR engines and large-scale real document image data (e.g., IIT-CDIP~\cite{iitcdip}) for a model pre-training.
Although they showed remarkable advances in recent years, extra effort is required to ensure the performance of an entire VDU model by using the off-the-shelf OCR engine.

%% file: tex/3.method.tex
\begin{figure*}[t!]
    \centering
    \includegraphics[width=\linewidth]{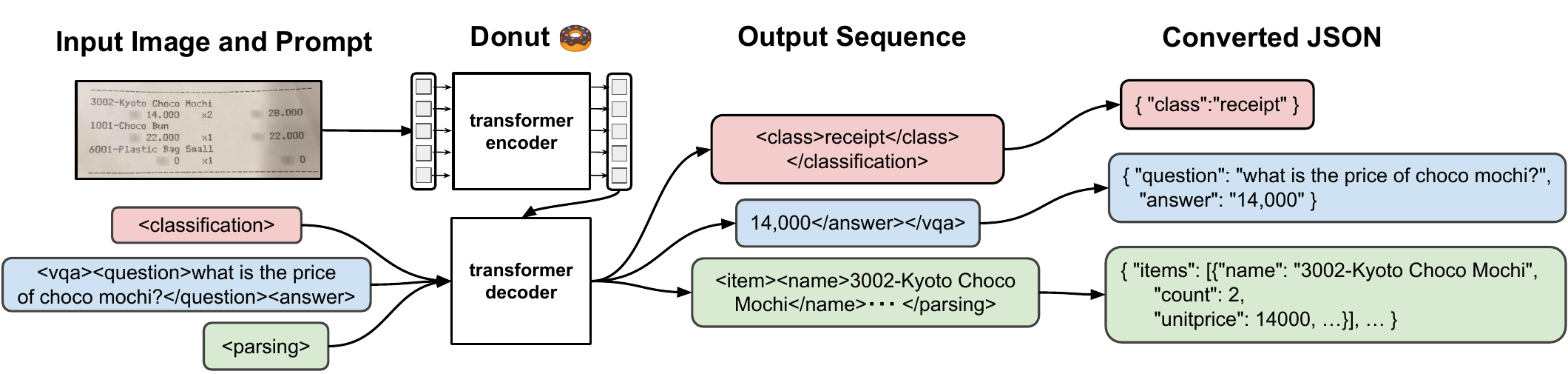}
    \caption{{\bf The pipeline of \oursb.} The encoder maps a given document image into embeddings. With the encoded embeddings, the decoder generates a sequence of tokens that can be converted into a target type of information in a structured form} %
    \label{fig:teaser}
\end{figure*}

\subsection{Document Understanding Transformer}
\ours is an end-to-end (i.e., self-contained) VDU model for general understanding of document images. %
The architecture of \ours is quite simple, which consists of a Transformer~\cite{vaswani2017transformer,dosovitskiy2020vit}-based visual encoder and textual decoder modules. Note that \ours does not rely on any modules related to OCR functionality but uses a visual encoder for extracting features from a given document image. The following textual decoder maps the derived features into a sequence of subword tokens to construct a desired structured format (e.g., JSON). Each model component is Transformer-based, and thus the model is trained easily in an end-to-end manner. The overall process of \ours is illustrated in Figure~\ref{fig:teaser}.

\subsubsection{Encoder.}

The visual encoder converts the input document image $\mathbf{x}{\in}\mathbb{R}^{H\times W\times C}$ into a set of embeddings $\{\mathbf{z}_{i} | \mathbf{z}_{i}{\in}\mathbb{R}^{d}, 1{\le}i{\le}n\}$, where $n$ is feature map size or the number of image patches and  $d$ is the dimension of the latent vectors of the encoder.
Note that CNN-based models~\cite{HeZRS16} or Transformer-based models~\cite{dosovitskiy2020vit,Liu_2021_ICCV} can be used as the encoder network. In this study, we use Swin Transformer~\cite{Liu_2021_ICCV} because it shows the best performance in our preliminary study in document parsing. %
Swin Transformer first splits the input image $\mathbf{x}$ into non-overlapping patches.
Swin Transformer blocks, consist of a shifted window-based multi-head self-attention module and a two-layer MLP, are applied to the patches.
Then, patch merging layers are applied to the patch tokens at each stage.
The output of the final Swin Transformer block $\{\mathbf{z}\}$ is fed into the following textual decoder.

\subsubsection{Decoder.}
Given the $\{\mathbf{z}\}$, the textual decoder generates a token sequence $(\mathbf{y}_{i})_{i=1}^{m}$, where $ \mathbf{y}_{i}{\in}\mathbb{R}^{v}$ is an one-hot vector for the $i$-th token, $v$ is the size of token vocabulary, and $m$ is a hyperparameter, respectively.
We use BART~\cite{lewis-etal-2020-bart} as the decoder architecture.
Specifically, we initialize the decoder model weights with those from the publicly available\footnote{\url{https://huggingface.co/hyunwoongko/asian-bart-ecjk}.} pre-trained multi-lingual BART model\cite{liu-etal-2020}.

\subsubsection{Model Input.}
Following the original Transformer~\cite{vaswani2017transformer}, we use a teacher-forcing scheme~\cite{williams1989learning}, which is a model training strategy that uses the ground truth as input instead of model output from a previous time step.
In the test phase, inspired by GPT-3~\cite{NEURIPS2020_1457c0d6}, the model generates a token sequence given a prompt.
We add new special tokens for the prompt for each downstream task in our experiments.
The prompts that we use for our applications are shown with the desired output sequences in Figure~\ref{fig:teaser}.
Illustrative explanations for the teacher-forcing strategy and the decoder output format are available in Appendix~\ref{sec:detail_of_scheme_and_format}. %

\subsubsection{Output Conversion.}
The output token sequence is converted to a desired structured format.
We adopt a JSON format due to its high representation capacity.
As shown in Figure~\ref{fig:teaser}, a token sequence is one-to-one invertible to a JSON data.
We simply add two special tokens \texttt{\small [START\_$\ast$]} and \texttt{\small [END\_$\ast$]}, where $\ast$ indicates each field to extract.
If the output token sequence is wrongly structured, we simply treat the field is lost. For example, if there is only \texttt{\small [START\_name]} exists but no \texttt{\small [END\_name]}, we assume the model fails to extract ``name'' field.
This algorithm can easily be implemented with simple regular expressions~\cite{Friedl06}.

\subsection{Pre-training} %

\subsubsection{Task.}\label{sec:pretraining}
The model is trained to read all texts in the image in reading order (from top-left to bottom-right, basically).
The objective is to minimize cross-entropy loss of next token prediction by jointly conditioning on the image and previous contexts.
This task can be interpreted as a pseudo-OCR task.
The model is trained as a visual language model over the visual corpora, i.e., document images.

\subsubsection{Visual Corpora.}
We use IIT-CDIP~\cite{iitcdip}, which is a set of 11M scanned english document images. A commercial CLOVA OCR API is applied to get the pseudo text labels.
As aforementioned, however, this kind of dataset is not always available, especially for languages other than English.
To alleviate the dependencies, we build a scalable \textit{\textbf{Synth}etic \textbf{Do}cument \textbf{G}enerator}, referred to as \textbf{SynthDoG}.
Using the SynthDog and Chinese, Japanese, Korean and English Wikipedia, we generated 0.5M samples per language.

\begin{figure}[t!]
    \centering
     \includegraphics[width=0.7\linewidth]{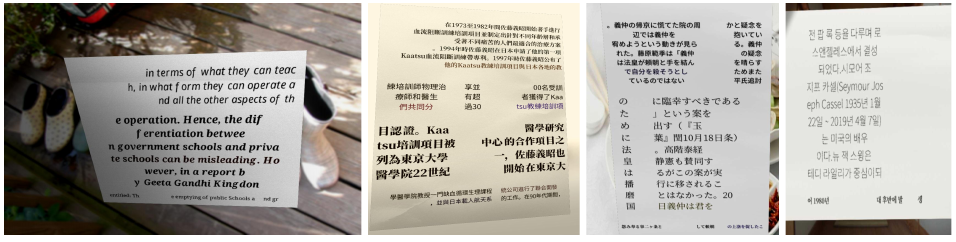}
    \caption{{\bf Generated English, Chinese, Japanese, and Korean samples with \textbf{SynthDoG}.} Heuristic random patterns are applied to mimic the real documents}%
    \label{fig:synthdog}
\end{figure}

\subsubsection{Synthetic Document Generator.}
The pipeline of image rendering basically follows Yim et al. \cite{synthtiger}.
As shown in Figure~\ref{fig:synthdog}, the generated sample consists of several components; background, document, text, and layout.
Background image is sampled from ImageNet~\cite{deng2009imagenet}, and a texture of document is sampled from the collected paper photos.
Words and phrases are sampled from Wikipedia.
Layout is generated by a simple rule-based algorithm that randomly stacks grids.
In addition, several image rendering techniques~\cite{Gupta16,long2020unrealtext,synthtiger} are applied to mimic real documents.
The generated examples are shown in Figure~\ref{fig:synthdog}.
More details of SynthDoG are available in the code\footref{footnote_code} and Appendix~\ref{sec:detail_of_synthdog}. %

\subsection{Fine-tuning}
After the model learns \textit{how to read}, in the application stage (i.e., fine-tuning), we teach the model \textit{how to understand} the document image.
As shown in Figure~\ref{fig:teaser}, we interpret all downstream tasks as a JSON prediction problem.

The decoder is trained to generate a token sequence that can be converted into a JSON that represents the desired output information. For example, in the document classification task, the decoder is trained to generate a token sequence \texttt{\small [START\_class][memo][END\_class]} which is 1-to-1 invertible to a JSON \{``class'': ``memo''\}.
We introduce some special tokens (e.g., \texttt{\small [memo]} is used for representing the class ``memo''), if such replacement is available in the target task.

%% file: tex/4.experiments.tex
\begin{figure*}[t!]
    \centering
    \includegraphics[width=\linewidth]{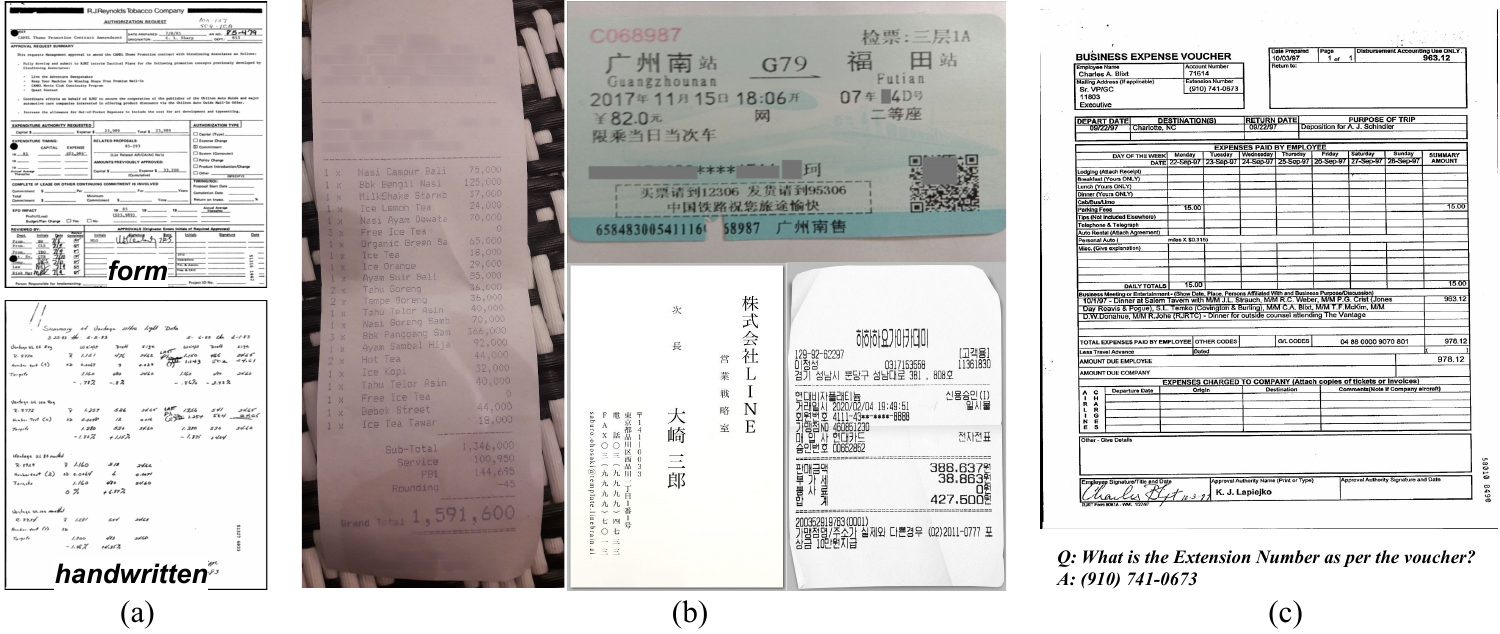}
    \caption{{\bf Samples of the downstream datasets.} (a) Document Classification. (b) Document Information Extraction. (c) Document Visual Question Answering} %
    \label{fig:datasets}
\end{figure*}

In this section, we present \oursb fine-tuning results on three VDU applications on six different datasets including both public benchmarks and private industrial service datasets. The samples are shown in Figure~\ref{fig:datasets}. %

\subsection{Downstream Tasks and Datasets}
\subsubsection{Document Classification.}
To see whether the model can distinguish across different types of documents, we test a classification task.
Unlike other models that predict the class label via a softmax on the encoded embedding, \ours generate a JSON that contains class information to maintain the uniformity of the task-solving method.
We report overall classification accuracy on a test set.

\paragraph{RVL-CDIP.}
The RVL-CDIP dataset~\cite{harley2015icdar} consists of 400K images in 16 classes, with 25K images per class. The classes include letter, memo, email, and so on. There are 320K training, 40K validation, and 40K test images. 

\subsubsection{Document Information Extraction.}
To see the model fully understands the complex layouts and contexts in documents, we test document information extraction (IE) tasks on various real document images including both public benchmarks and real industrial datasets.
In this task, the model aims to map each document to a structured form of information that is consistent with the target ontology or database schema. See Figure~\ref{fig:problem_definition} for an illustrative example. The model should not only read the characters well, but also understand the layouts and semantics to infer the groups and nested hierarchies among the texts.

We evaluate the models with two metrics; field-level F1 score~\cite{hwang2019pot,xu2019_layoutLM,hong2021bros} and Tree Edit Distance (TED) based accuracy~\cite{ted,teds,hwang2021costeffective}.
The F1 checks whether the extracted field information is in the ground truth. Even if a single character is missed, the score assumes the field extraction is failed.
Although F1 is simple and easy to understand, there are some limitations. First, it does not take into account partial overlaps. Second, it can not measure the predicted structure (e.g., groups and nested hierarchy).
To assess overall accuracy, we also use another metric based on TED~\cite{ted}, that can be used for any documents represented as trees. It is calculated as, $\max(0, 1-\text{TED}(\text{pr},\text{gt})/\text{TED}(\phi,\text{gt}))$, where $\text{gt}$, $\text{pr}$, and $\phi$ stands for ground truth, predicted, and empty trees respectively. Similar metrics are used in recent works on document IE~\cite{teds,hwang2021costeffective}

We use two public benchmark datasets as well as two private industrial datasets which are from our active real-world service products.
Each dataset is explained in the followings.

\paragraph{CORD.}
The Consolidated Receipt Dataset (CORD)\footnote{\url{https://huggingface.co/datasets/naver-clova-ix/cord-v1}.}\cite{park2019cord} is a public benchmark that consists of 0.8K train, 0.1K valid, 0.1K test receipt images.
The letters of receipts is in Latin alphabet.
The number of unique fields is 30 containing menu name, count, total price, and so on.
There are complex structures (i.e., nested groups and hierarchies such as \texttt{\small items>item>{\scriptsize\{}name, count, price{\scriptsize\}}}) in the information.
See Figure~\ref{fig:problem_definition} for more details.

\paragraph{Ticket.}
This is a public benchmark dataset~\cite{eaten} that consists of 1.5K train and 0.4K test Chinese train ticket images. 
We split 10\% of the train set as a validation set.
There are 8 fields which are ticket number, starting station, train number, and so on.
The structure of information is simple and all keys are guaranteed to appear only once and the location of each field is fixed.

\paragraph{Business Card (In-Service Data).}
This dataset is from our active products that are currently deployed.
The dataset consists of 20K train, 0.3K valid, 0.3K test Japanese business cards.
The number of fields is 11, including name, company, address, and so on.
The structure of information is similar to the \textit{Ticket} dataset.

\paragraph{Receipt (In-Service Data).}
This dataset is also from one of our real products. The dataset consists of 40K train, 1K valid, 1K test Korean receipt images.
The number of unique field is 81, which includes store information, payment information, price information, and so on.
Each sample has complex structures compared to the aforementioned datasets.
Due to industrial policies, not all samples can publicly be available. Some real-like high-quality samples are shown in Figure~\ref{fig:datasets} and in the supplementary material.

\subsubsection{Document Visual Question Answering.}
To validate the further capacity of the model, we conduct a document visual question answering task (DocVQA).
In this task, a document image and question pair is given and the model predicts the answer for the question by capturing both visual and textual information within the image. %
We make the decoder generate the answer by setting the question as a starting prompt to keep the uniformity of the method (See Figure~\ref{fig:teaser}).

\paragraph{DocVQA.}
The dataset is from Document Visual Question Answering competition\footnote{\url{https://rrc.cvc.uab.es/?ch=17}.} and consists of 50K questions defined on more than 12K documents~\cite{mathew2021docvqa}.
There are 40K train, 5K valid, and 5K test questions.
The evaluation metric is ANLS (Average Normalized Levenshtein Similarity) which is an edit-distance-based metric.
The score on the test set is measured via the evaluation site.

\subsection{Setups}
We use Swin-B~\cite{Liu_2021_ICCV} as a visual encoder of \oursb with slight modification.
We set the layer numbers and window size as $\{2, 2, 14, 2\}$ and 10.
In further consideration of the speed-accuracy trade-off, we use the first four layers of BART as a decoder.
As explained in Section~\ref{sec:pretraining}, we train the multi-lingual \oursb using the 2M synthetic and 11M IIT-CDIP scanned document images.
We pre-train the model for 200K steps with 64 A100 GPUs and a mini-batch size of 196.
We use Adam~\cite{Adamoptim} optimizer, the learning rate is scheduled and the initial rate is selected from 1e-5 to 1e-4.
The input resolution is set to 2560$\times$1920 and a max length in the decoder is set to 1536.
All fine-tuning results are achieved by starting from the pre-trained multi-lingual model.
Some hyperparameters are adjusted at fine-tuning and in ablation studies.
We use 960$\times$1280 for Train Tickets and Business Card parsing tasks.
We fine-tune the model while monitoring the edit distance over token sequences.
The speed of \ours is measured on a P40 GPU, which is much slower than A100.
For the OCR based baselines, states-of-the-art OCR engines are used, including MS OCR API used in \cite{xu-etal-2021-layoutlmv2} and CLOVA OCR API\footnote{\url{https://clova.ai/ocr}.} used in \cite{hwang2020spade,hwang2021costeffective}.
An analysis on OCR engines is available in Section~\ref{sec:ablation_and_analysis}.
More details of OCR and training setups are available in Appendix~\ref{sec:detail_of_ocr_engines} and \ref{sec:detail_of_implementation_and_hyperparams}. %

\begin{table*}[t]
\centering
\caption{{\bf Classification results on the RVL-CDIP dataset.} \oursb achieves state-of-the-are performance with reasonable speed and model size efficiency. \oursb is a general purpose backbone but does not rely on OCR while other recent backbones (e.g., LayoutLM) do. $^\dag$\# parameters for OCR should be considered for non-E2E models} \label{tbl:docclass}
\begin{threeparttable}
    \centering
  \begin{tabular}{lcccc}
  \toprule
   & OCR  & \#Params & Time (ms) & Accuracy (\%)\\
    \midrule
    BERT &\checkmark & 110M + $\alpha^{\dag}$ & 1392 & 89.81 \\ %
    RoBERTa &\checkmark & 125M + $\alpha^{\dag}$  & 1392 & 90.06 \\ %
    LayoutLM &\checkmark & 113M + $\alpha^{\dag}$  & 1396 & 91.78 \\
    LayoutLM (w/ image) &\checkmark & 160M + $\alpha^{\dag}$  & 1426 & 94.42 \\
    LayoutLMv2 &\checkmark & 200M + $\alpha^{\dag}$  & 1489 & {95.25} \\ %
    \midrule
    \oursb \textbf{(Proposed)}&  & 143M & \textbf{752} & \textbf{95.30} \\
    \bottomrule
  \end{tabular}
\end{threeparttable}
\end{table*}

\subsection{Experimental Results}
\subsubsection{Document Classification.}
The results are shown in Table~\ref{tbl:docclass}.
Without relying on any other resource (e.g., off-the-shelf OCR engine), \ours shows a state-of-the-art performance among the general-purpose VDU models such as LayoutLM~\cite{xu2019_layoutLM} and LayoutLMv2~\cite{xu-etal-2021-layoutlmv2}. 
In particular, \ours surpasses the LayoutLMv2 accuracy reported in \cite{xu-etal-2021-layoutlmv2}, while using fewer parameters with the 2x faster speed. %
Note that the OCR-based models must consider additional model parameters and speed for the entire OCR framework, which is not small in general. For example, a recent advanced OCR-based model~\cite{baek2019craft,baek2019wrong} requires more than 80M parameters. Also, training and maintaining the OCR-based systems are costly~\cite{hwang2021costeffective}, leading to needs for the \ours-like end-to-end approach.

\begin{table*}[t]
\centering
\caption{{\bf Performances on various document IE tasks.} The field-level F1 scores and tree-edit-distance-based accuracies are reported. \oursb shows the best accuracies for all domains with significantly faster inference speed. $^{\dag}$Parameters for vocabulary are omitted for fair comparisons among multi-lingual models. $^\ddag$\# parameters for OCR should be considered. $^{\ast}$Official multi-lingual extension models are used} \label{tbl:information_extraction} %
\begin{adjustbox}{max width=\textwidth}
\begin{threeparttable}
  \centering
  \begin{tabular}{lcccccccccccccc}
  \toprule
   &&& \multicolumn{3}{c}{CORD~\cite{park2019cord}}& \multicolumn{3}{c}{Ticket~\cite{eaten}} & \multicolumn{3}{c}{Business Card} & \multicolumn{3}{c}{Receipt}\\
   \cmidrule(lr){4-6}\cmidrule(lr){7-9}\cmidrule(lr){10-12}\cmidrule(lr){13-15}
    & OCR & \#Params & {\scriptsize Time (s)} & F1 & Acc. & {\scriptsize Time (s)} & F1 & Acc. & {\scriptsize Time (s)} & F1 & Acc. & {\scriptsize Time (s)} & F1 & Acc.\\
    \midrule
    BERT$^{\ast}$~\cite{hwang2019pot} &\checkmark& $86^{\dag}_{\text{M}}+\alpha^{\ddag}$&
    1.6&73.0&65.5&  1.7&74.3&82.4& 1.5&40.8&72.1& 2.5&70.3&54.1\\
    BROS~\cite{hong2021bros} &\checkmark& $86^{\dag}_{\text{M}}+\alpha^{\ddag}$&
    1.7&74.7&70.0&  \multicolumn{9}{r}{}\\
    LayoutLM~\cite{xu2019_layoutLM} &\checkmark& $89^{\dag}_{\text{M}}+\alpha^{\ddag}$&
    1.7&78.4&81.3&  \multicolumn{9}{r}{}\\
    LayoutLMv2$^{\ast}$~\cite{xu-etal-2021-layoutlmv2,layoutxlm} &\checkmark& $179^{\dag}_{\text{M}}+\alpha^{\ddag}$&
    1.7&78.9&82.4&  1.8&87.2&90.1& 1.6&52.2&83.0& 2.6&72.9&78.0\\
    \midrule
    \oursb &  &  $143^{\dag}_{\text{M}}$&
    \textbf{1.2}&\textbf{84.1}&\textbf{90.9}&\textbf{0.6}&\textbf{94.1}&\textbf{98.7}&\textbf{1.4}&\textbf{57.8}&\textbf{84.4}&\textbf{1.9}&\textbf{78.6}&\textbf{88.6}\\
    \midrule
    \midrule
    SPADE$^{\ast}$~\cite{hwang-etal-2021-spatial}  &\checkmark& $93^{\dag}_{\text{M}} + \alpha^{\ddag}$&
    4.0&74.0&75.8&  4.5&14.9&29.4& 4.3&32.3&51.3& 7.3&64.1&53.2\\
    WYVERN$^{\ast}$~\cite{hwang-etal-2020-towards}  &\checkmark& $106^{\dag}_{\text{M}} + \alpha^{\ddag}$&
    1.2&43.3&46.9&  1.5&41.8&54.8& 1.7&29.9&51.5& 3.4&71.5&82.9\\
    \bottomrule
  \end{tabular}
  \vspace{-0.6em}
\end{threeparttable}
\end{adjustbox}
\end{table*}

\subsubsection{Document Information Extraction.}

Table~\ref{tbl:information_extraction} shows the results on the four different document IE tasks.
The first group uses a conventional BIO-tagging-based IE approach~\cite{hwang2019pot}. We follows the conventions in IE~\cite{xu2019_layoutLM,hong2021bros}. OCR extracts texts and bounding boxes from the image, and then the serialization module sorts all texts with geometry information within the bounding box. The BIO-tagging-based named entity recognition task performs token-level tag classification upon the ordered texts to generate a structured form. We test three general-purpose VDU backbones, BERT~\cite{devlinBERT2018}, BROS~\cite{hong2021bros}, LayoutLM~\cite{xu2019_layoutLM}, and LayoutLMv2~\cite{xu-etal-2021-layoutlmv2,layoutxlm}.

We also test two recently proposed IE models, SPADE~\cite{hwang2020spade} and WYVERN~\cite{hwang2021costeffective}. SPADE is a graph-based IE method that predicts relations between bounding boxes. WYVERN is an Transformer encoder-decoder model that directly generates entities with structure given OCR outputs. WYVERN is different from \ours in that it takes the OCR output as its inputs.

For all domains, including public and private in-service datasets, \ours shows the best scores among the comparing models.
By measuring both F1 and TED-based accuracy, we observe not only \ours can extract key information but also predict complex structures among the field information.
We observe that a large input resolution gives robust accuracies but makes the model slower. For example, the performance on the CORD with 1280$\times$960 was 0.7 sec./image and 91.1 accuracy. But, the large resolution showed better performances on the low-resource situation. The detailed analyses are in Section~\ref{sec:ablation_and_analysis}.
Unlike other baselines, \ours shows stable performance regardless of the size of datasets and complexity of the tasks (See Figure~\ref{fig:datasets}). This is a significant impact as the target tasks are already actively used in industries.

\subsubsection{Document Visual Question Answering.}
Table~\ref{tbl:docvqa} shows the results on the DocVQA dataset.
The first group is the general-purposed VDU backbones whose scores are from the LayoutLMv2 paper~\cite{xu-etal-2021-layoutlmv2}.
We measure the running time with MS OCR API used in \cite{xu-etal-2021-layoutlmv2}.
The model in the third group is a DocVQA-specific-purposed fine-tuning model of LayoutLMv2, 
whose inference results are available in the official leader-board.\footnote{\url{https://rrc.cvc.uab.es/?ch=17&com=evaluation&task=1}.}

\begin{figure}[t]
    \centering
    \includegraphics[width=\linewidth]{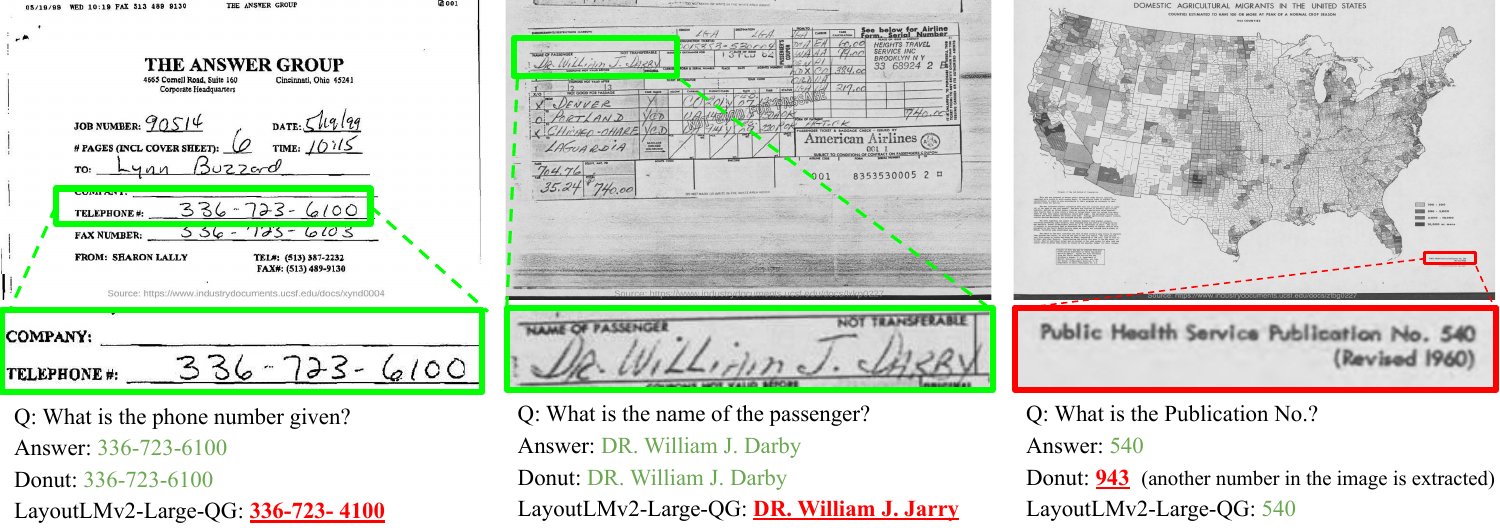}
    \caption{{\bf Examples of Donut and LayoutLMv2 outputs on DocVQA.} The OCR-errors make a performance upper-bound of the OCR-dependent baselines, e.g., LayoutLMv2 (left and middle examples). Due to the input resolution constraint of the end-to-end pipeline, Donut miss some tiny texts in large-scale images (right example) but this could be mitigated by scaling the input image size (See Section~\ref{sec:ablation_and_analysis})}
    \label{fig:doc_vqa_example}
\end{figure}

As can be seen, \oursb achieves competitive scores with the baselines that are dependent on external OCR engines. Especially, \ours shows that it is robust to the handwritten documents, which is known to be challenging to process. In the conventional approach, adding a post-processing module that corrects OCR errors is an option to strengthen the pipeline~\cite{schaefer-neudecker-2020-two,rijhwani-etal-2020-ocr,duong-etal-2021-unsupervised} or adopting an encoder-decoder architecture on the OCR outputs can mitigate the problems of OCR errors~\cite{hwang2021costeffective}. However, this kind of approaches tend to increase the entire system size and maintenance cost. \ours shows a completely different direction.
Some inference results are shown in Figure~\ref{fig:doc_vqa_example}. The samples show the current strengths of \ours as well as the left challenges in the \ours-like end-to-end approach. Further analysis and ablation is available in Section~\ref{sec:ablation_and_analysis}.

\begin{table}[t]
\centering
\caption{{\bf Average Normalized Levenshtein Similarity (ANLS) scores on DocVQA.} \oursb shows a promising result without OCR. $^{\ast}$\oursb shows a high ANLS score on the handwritten documents which are known to be challenging due to the difficulty of handwriting OCR (See Figure~\ref{fig:doc_vqa_example}). $^\dag$Token embeddings for English is counted for a fair comparison. $^\ddag$\# parameters for OCR should be considered}  \label{tbl:docvqa} %
\begin{adjustbox}{max width=\linewidth}
\begin{threeparttable}
\footnotesize
  \centering
  \begin{tabular}{lcccccc}
  \toprule %
  & Fine-tuning set & OCR & \#Params$^{\dag}$ & Time (ms) & $^{\text{ANLS}^{\:}}_{\text{test set}}$ & $^{\text{ANLS}^\ast}_{\text{handwritten}}$\\
    \midrule %
    BERT \cite{xu-etal-2021-layoutlmv2}&train set&\checkmark & 110M + $\alpha^{\ddag}$ & 1517 & 63.5&n/a\\
    LayoutLM\cite{xu2019_layoutLM}&train set&\checkmark & 113M + $\alpha^{\ddag}$ & 1519 & 69.8 &n/a\\
    LayoutLMv2\cite{xu-etal-2021-layoutlmv2}&train set&\checkmark & 200M + $\alpha^{\ddag}$ & 1610 & 78.1 &n/a\\ %
    \midrule
    \oursb &train set& & 176M & \textbf{782} & 67.5 &\textbf{72.1}\\ 
    \midrule
    \midrule
    LayoutLMv2-Large-QG\cite{xu-etal-2021-layoutlmv2}&train + dev + QG&\checkmark & 390M + $\alpha^{\ddag}$ & 1698 & \textbf{86.7} &67.3\\
    \bottomrule
  \end{tabular}
\end{threeparttable}
\end{adjustbox}
\end{table}

\input{tex/4.expriments2}

%% file: tex/4.expriments2.tex
\subsection{Further Studies}
\label{sec:ablation_and_analysis}

\begin{figure}[t]
    \centering
    \includegraphics[width=\linewidth]{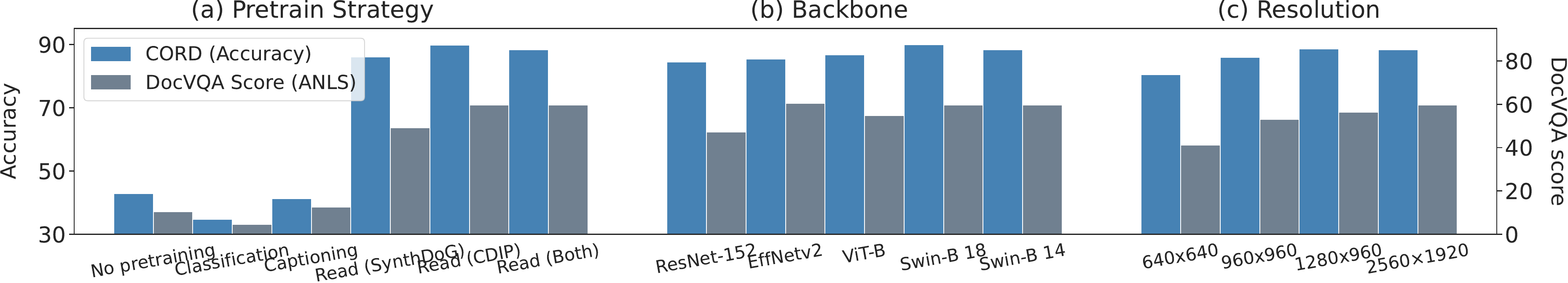}
    \caption{\textbf{Analysis on (a) pre-training strategies, (b) image backbones, and (c) input resolutions.} Performances on CORD~\cite{park2019cord} and DocVQA~\cite{mathew2021docvqa} are shown} %
    \label{fig:analysis_ablation}
\end{figure}

In this section, we study several elements of understanding \ours. We show some striking characteristics of \ours through the experiments and visualization.

\subsubsection{On Pre-training Strategy.}
We test several pre-training tasks for VDUs. Figure~\ref{fig:analysis_ablation}(a) shows that the \ours pre-training task (i.e., text reading) is the most simple yet effective approach. Other tasks that impose a general knowledge of images and texts on models, e.g., image captioning, show little gains in the fine-tuning tasks.
For the text reading tasks, we verify three options, SynthDoG only, IIT-CDIP only, and both.
Note that synthetic images were enough for the document IE task in our analysis. 
However, in the DocVQA task, it was important to see the real images. 
This is probably because the image distributions of IIT-CDIP and DocVQA are similar~\cite{mathew2021docvqa}.

\subsubsection{On Encoder Backbone.}
Here, we study popular image classification backbones that show superior performance in traditional vision tasks to measure their performance in VDU tasks.
The Figure~\ref{fig:analysis_ablation}(b) shows the comparison results. We use all the backbones pre-trained on ImageNet~\cite{deng2009imagenet}.
EfficientNetV2~\cite{tan2021efficientnetv2} and Swin Transformer~\cite{Liu_2021_ICCV} outperform others on both datasets. We argue that this is due to the high expressiveness of the backbones, which were shown by the striking scores on several downstream tasks as well.
We choose Swin Transformer due to the high scalability of the Transformer-based architecture and higher performance over the EfficientNetV2's.

\subsubsection{On Input Resolution.}
The Figure~\ref{fig:analysis_ablation}(c) shows the performance of \ours grows rapidly as we set a larger input size. This gets clearer in the DocVQA where the images are larger with many tiny texts. %
But, increasing the size for a precise result incurs bigger computational costs. Using an efficient attention mechanism~\cite{wang2020linformer} may avoid the matter in architectural design, but we use the original Transformer~\cite{vaswani2017transformer} as we aim to present a simpler architecture in this work.

\subsubsection{On Text Localization.}
\begin{figure}[t]
    \centering
    \includegraphics[width=\linewidth]{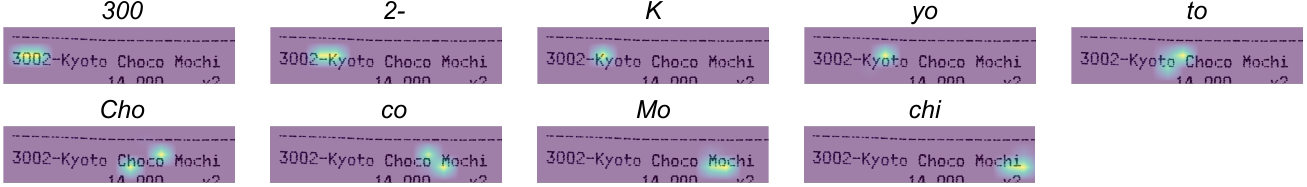}
    \caption{{\bf Visualization of cross-attention maps in the decoder and its application to text localization.} \ours is trained without any supervision for the localization. The \ours decoder attends proper text regions to process the image}
    \label{fig:detection}
\end{figure}

To see how the model behaves, we visualize the corss attention maps of the decoder given an unseen document image. As can be seen in Figure~\ref{fig:detection}, the model shows meaningful results that can be used as an auxiliary indicator. The model attends to a desired location in the given image.

\begin{figure}[!t]
    \centering
    \includegraphics[width=\linewidth]{figures/fig9.pdf}
    \caption{\textbf{Comparison of BERT, LayoutLMv2 and \ours on CORD.} The performances (i.e., speed and accuracy) of the OCR-based models extremely varies depending on what OCR engine is used (left and center). \ours shows robust performances even in a low resourced situation showing the higher score only with 80 samples (right)}
    \label{fig:analysis_ablation_ocr_low_resource}
\end{figure}
\subsubsection{On OCR System.}
We test four widely-used public OCR engines (See Figure~\ref{fig:analysis_ablation_ocr_low_resource}). The results show that the performances (i.e., speed and accuracy) of the conventional OCR-based methods heavily rely on the off-the-shelf OCR engine.
More details of the OCR engines are available in Appendix~\ref{sec:detail_of_ocr_engines}. %

\subsubsection{On Low Resourced Situation.}
We evaluate the models by limiting the size of training set of CORD~\cite{park2019cord}. The performance curves are shown in the right Figure~\ref{fig:analysis_ablation_ocr_low_resource}. \ours shows a robust performances. We also observe that a larger input resolution, 2560$\times$1920, shows more robust scores on the extremely low-resourced situation, e.g., 80 samples. As can be seen, \ours outperformed the LayoutLMv2 accuracy only with 10\% of the data, which is only 80 samples.

%% file: tex/5.related.tex
\subsection{Optical Character Recognition}
Recent trends of OCR study are to utilize deep learning models in its two sub-steps: 1) text areas are predicted by a detector; 2) a text recognizer then recognizes all characters in the cropped image instances. Both are trained with a large-scale datasets including the synthetic images~\cite{Jaderberg14c,Gupta16} and real images~\cite{7333942,Phan_2013_ICCV}. %

Early detection methods used CNNs to predict local segments and apply heuristics to merge them~\cite{Huang10.1007/978-3-319-10593-2_33,Zhang_2016_CVPR}.
Later, region proposal and bounding box regression based methods were proposed~\cite{LiaoSBWL17}. %
Recently, focusing on the homogeneity and locality of texts, component-level approaches were proposed~\cite{CTPN,baek2019craft}. 

Many modern text recognizer share a similar approach~\cite{starnet,Shi2016RobustST,Shi2017AnET,jianfeng2017deep} that can be interpreted into a combination of several common deep modules~\cite{baek2019wrong}. %
Given the cropped text instance image,
most recent text recognition models apply CNNs to encode the image into a feature space. A decoder is then applied to extract characters from the features.

\subsection{Visual Document Understanding}
Classification of the document type is a core step towards automated document processing.
Early methods treated the problem as a general image classification, so various CNNs were tested~\cite{Kang2014ConvolutionalNN,7333933,7333910}. Recently, with  BERT~\cite{devlinBERT2018}, the methods based on a combination of CV and NLP were widely proposed~\cite{xu2019_layoutLM,li-etal-2021-structurallm}.
As a common approach, most methods rely on an OCR engine to extract texts; then the OCR-ed texts are serialized into a token sequence; finally they are fed into a language model (e.g., BERT) with some visual features if available.
Although the idea is simple, the methods showed remarkable performance improvements and became a main trend in recent years~\cite{xu-etal-2021-layoutlmv2,selfdoc,Appalaraju_2021_ICCV}. %

Document IE covers a wide range of real applications~\cite{hwang2019pot,majumder2020representation}, for example, given a bunch of raw receipt images, a document parser can automate a major part of receipt digitization, which has been required numerous human-labors in the traditional pipeline.
Most recent models~\cite{hwang-etal-2021-spatial,hwang2021costeffective} take the output of OCR as their input.
The OCR results are then converted to the final parse through several processes, which are often complex.
Despite the needs in the industry, only a few works have been attempted on end-to-end parsing. Recently, some works are proposed to simplify the complex parsing processes~\cite{hwang-etal-2021-spatial,hwang2021costeffective}.
But they still rely on a separate OCR to extract text information.

Visual QA on documents seeks to answer questions asked on document images. This task requires reasoning over visual elements of the image and general knowledge to infer the correct answer~\cite{mathew2021docvqa}. Currently, most state-of-the-arts follow a simple pipeline consisting of applying OCR followed by BERT-like transformers~\cite{xu2019_layoutLM,xu-etal-2021-layoutlmv2}. %
However, the methods work in an extractive manner by their nature. Hence, there are some concerns for the question whose answer does not appear in the given image~\cite{icdar21docvqa}. To tackle the concerns, generation-based methods have also been proposed~\cite{10.1007/978-3-030-86331-9_47}.

%% file: tex/6.conclusion.tex
In this work, we propose a novel end-to-end framework for visual document understanding. The proposed method, \oursb, directly maps an input document image into a desired structured output.
Unlike conventional methods, \oursb does not depend on OCR and can easily be trained in an end-to-end fashion.
We also propose a synthetic document image generator, {SynthDoG}, to alleviate the dependency on large-scale real document images and we show that \oursb can be easily extended to a multi-lingual setting.
We gradually trained the model from \textit{how to read} to \textit{how to understand} through the proposed training pipeline.
Our extensive experiments and analysis on both external public benchmarks and private internal service datasets show higher performance and better \textit{cost-effectiveness} of the proposed method. This is a significant impact as the target tasks are already practically used in industries.
Enhancing the pre-training objective could be a future work direction. We believe our work can easily be extended to other domains/tasks regarding document understanding.